\documentclass{article}
\usepackage{spconf,amsmath,graphicx}
\usepackage{amsfonts, multirow, booktabs}
\usepackage{threeparttable}
\usepackage{cite}
\usepackage{algorithm,algorithmic}
\usepackage[caption=false]{subfig}

\title{SINGLE DOMAIN DYNAMIC GENERALIZATION FOR IRIS PRESENTATION ATTACK DETECTION}
\name{Yachun Li\sthanks{Equal contribution.} , Jingjing Wang\footnotemark[1], Yuhui Chen, Di Xie\sthanks{Corresponding author (xiedi@hikvision.com).}, Shiliang Pu}
\address{Hikvision Research Institute}
\begin{document}
\newcommand{\F}{\mathbf{F}}
\newcommand{\D}{\mathbf{D}}
\newcommand{\C}{\mathbf{C}}

\ninept
\maketitle
\begin{abstract}
Iris presentation attack detection (PAD) has achieved great success under intra-domain settings but easily degrades on unseen domains. Conventional domain generalization methods mitigate the gap by learning domain-invariant features. However, they ignore the discriminative information in the domain-specific features. Moreover, we usually face a more realistic scenario with only one single domain available for training. To tackle the above issues, we propose a Single Domain Dynamic Generalization (SDDG) framework, which simultaneously exploits domain-invariant and domain-specific features on a per-sample basis and learns to generalize to various unseen domains with numerous natural images. Specifically, a dynamic block is designed to adaptively adjust the network with a dynamic adaptor. And an information maximization loss is further combined to increase diversity. The whole network is integrated into the meta-learning paradigm. We generate amplitude perturbed images and cover diverse domains with natural images. Therefore, the network can learn to generalize to the perturbed domains in the meta-test phase. Extensive experiments show the proposed method is effective and outperforms the state-of-the-art on LivDet-Iris 2017 dataset. 
\end{abstract}
\section{Introduction}
\label{sec:intro}
Iris recognition systems have been widely used due to the uniqueness and stability of iris patterns. However, the vulnerability to presentation attacks brings security threats to these systems. To tackle this issue, iris presentation attack detection (PAD) has been introduced to ensure the reliability of the recognition systems \cite{boyd2020iris}. 
\cite{he2016multi, raghavendra2017contlensnet, hoffman2018convolutional, fang2020micro} divide iris images into multiple patches to enhance patch-wise attack information, in order to exploit the intricate texture patterns of the iris. More recently, attention mechanism \cite{sharma2020d, fang2021iris, li2022few} has been introduced for a more interpretable focus on attack-related features. 

Despite the remarkable success, most iris PAD works focus on intra-dataset evaluations \cite{fang2021iris, li2022few}, which assume training and testing samples are from similar distributions. This assumption may not always hold in practical scenarios. The potential discrepancies are from capturing equipment, illuminations, background, etc. To solve this problem, \textit{domain adaptation} (DA) \cite{wang2021self, li2022few, zhou2022generative} and \textit{domain generalization} (DG) \cite{shao2020regularized, jia2020single, wang2022domain} techniques have been extensively studied in the PAD field. DA aims to minimize the domain discrepancy between a labeled source domain and an unlabeled target domain, while DG learns to generalize to an unknown target domain with multiple source domains. 
Conventional DG methods align features to a domain-agnostic space \cite{tzeng2014deep, ganin2016domain} so that the shared space can generalize well even on the unseen domains. For a more realistic scenario \textit{single domain generalization} (Single-DG), data generation is used to enlarge the distribution shifts and learn the domain-invariant representations \cite{li2021progressive, wang2021learning}. 
Although domain-invariant features are common among different domains, the neglected domain-specific features could still promote performance in each domain. Concretely, each domain, or each sample, has its unique characteristics and can be regarded as a latent space. Forced invariance among latent spaces strengthens the model's generalization ability on unseen domains, yet it inevitably discards discriminative information of individual latent spaces that could have facilitated visual tasks. 

In general, we usually have images from only one single domain in real-world iris PAD applications. To fully utilize the complementary information for generalized detection with a single domain, we propose a novel Single Domain Dynamic Generalization (SDDG) framework to dynamically extract the domain-invariant and domain-specific features. To the best of our knowledge, this is the first work to exploit dynamic generalization for Single-DG. 
The domain-invariant part is achieved by a common convolution followed by Instance Normalization (IN) \cite{ulyanov2017improved}, which is invariant to appearance changes \cite{pan2018two}. The domain-specific one is formed by multiple convolutions with a dynamic adaptor. The dynamic adaptor predicts the weights of each convolution in a sample-wise manner and is further combined with information maximization \cite{hu2017learning, liang2020we} to encourage disparate representations. 
Thus, our dynamic block is capable of adapting to the characteristics of each sample, and both domain-invariant and domain-specific features are used to improve the generalization.

Nevertheless, common DG paradigm requires the availability of data from multiple source domains, whereas Single-DG encounters a more realistic challenge with only one single source domain available \cite{qiao2020learning}. 
Although PAD samples from different domains are not available in Single-DG, we can easily get access to a large number of natural images from common datasets, such as ImageNet \cite{deng2009imagenet} and COCO \cite{lin2014microsoft}. These natural images cover adequate diverse scenes and can benefit the generalization ability. 

In order to best take advantage of diverse distributions from numerous natural images, we propose a meta-learning based paradigm and learn to generalize on the various unseen domains. We first conduct normal training based on source images in the meta-train phase. Then we learn to generalize to the perturbed images in the meta-test phase. Specifically, we perturb the original image based on Fourier transformation with natural images. Fourier transformation has the property that the phase component of Fourier spectrum preserves high-level semantics of the original signal, while the amplitude component contains low-level statistics \cite{oppenheim1981importance}. Following \cite{xu2021fourier}, we conduct the perturbation through Mixup \cite{zhang2017mixup} in the amplitude spectrum and generate diverse images. Moreover, we aim to adaptively learn domain-invariant and domain-specific representations on unseen domains and correctly adjust the sample-wise weights with the dynamic adaptor. Hence, instead of directly applying vanilla meta-learning for DG \cite{li2018learning} to learn domain-invariant representations, we conduct meta-optimization only in the dynamic block and focus on increasing the flexibility of the dynamic block under different distributions. 
Consequently, the domain-invariant and domain-specific features from the dynamic block are still plausible in Single-DG. 

Our main contributions are as follows: (1) We propose a novel SDDG framework to address single domain generalization for iris presentation attack detection by simultaneously exploiting domain-invariant and domain-specific features on a per-sample basis. (2) A meta-learning based paradigm is proposed to learn to generalize on the various domains perturbed with numerous natural images.  (3) Extensive experiments and visualizations on LivDet-Iris 2017 dataset demonstrate the effectiveness of the proposed method. 

\begin{figure}[t]
\begin{center}
\includegraphics[width=0.85\linewidth]{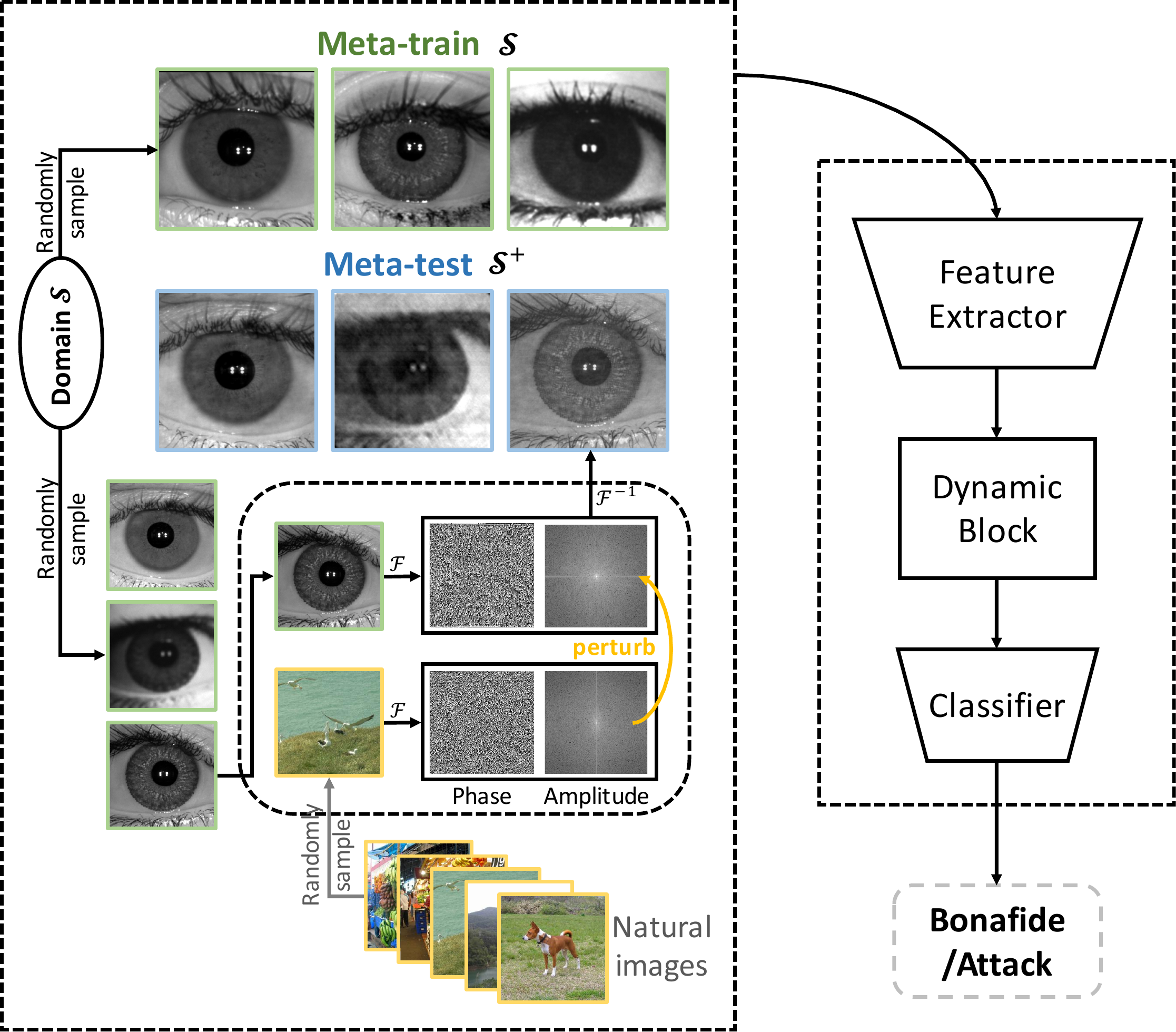}
\end{center}
\vspace{-0.5cm}
\caption{Overview of the proposed SDDG framework. }
\label{fig:overview}
\vspace{-0.3cm}
\end{figure}

\section{Proposed Method}
\label{sec:method}
Given a training set from a single domain, the proposed Single Domain Dynamic Generalization (SDDG) framework adapts network parameters on a per-sample basis and generalizes to diverse domains with natural images. 
To leverage the domain-specific features as the complementation to the domain-invariant ones, we design a novel dynamic block, which contains two branches. The domain-invariant branch diminishes instance style with Instance Normalization (IN) and reduces domain discrepancy. The domain-specific branch has multiple convolutions and utilizes the dynamic adaptor to adjust the sample-wise weights for its convolutions. 
The whole network is composed of a feature extractor (denoted as $\F$), a dynamic block (denoted as $\D$), and a classifier (denoted as $\C$), and it is integrated into the meta-learning paradigm. We generate the amplitude perturbed images through Fourier transformation and cover diverse domains with natural images. The various domains seen in the meta-test phase facilitate the generalization ability of the network. 
The overall SDDG framework is illustrated in Fig. \ref{fig:overview}. 

\subsection{Dynamic Block}
\label{ssec:dynamic}
\begin{figure}[t!]
\begin{center}
\includegraphics[width=0.85\linewidth]{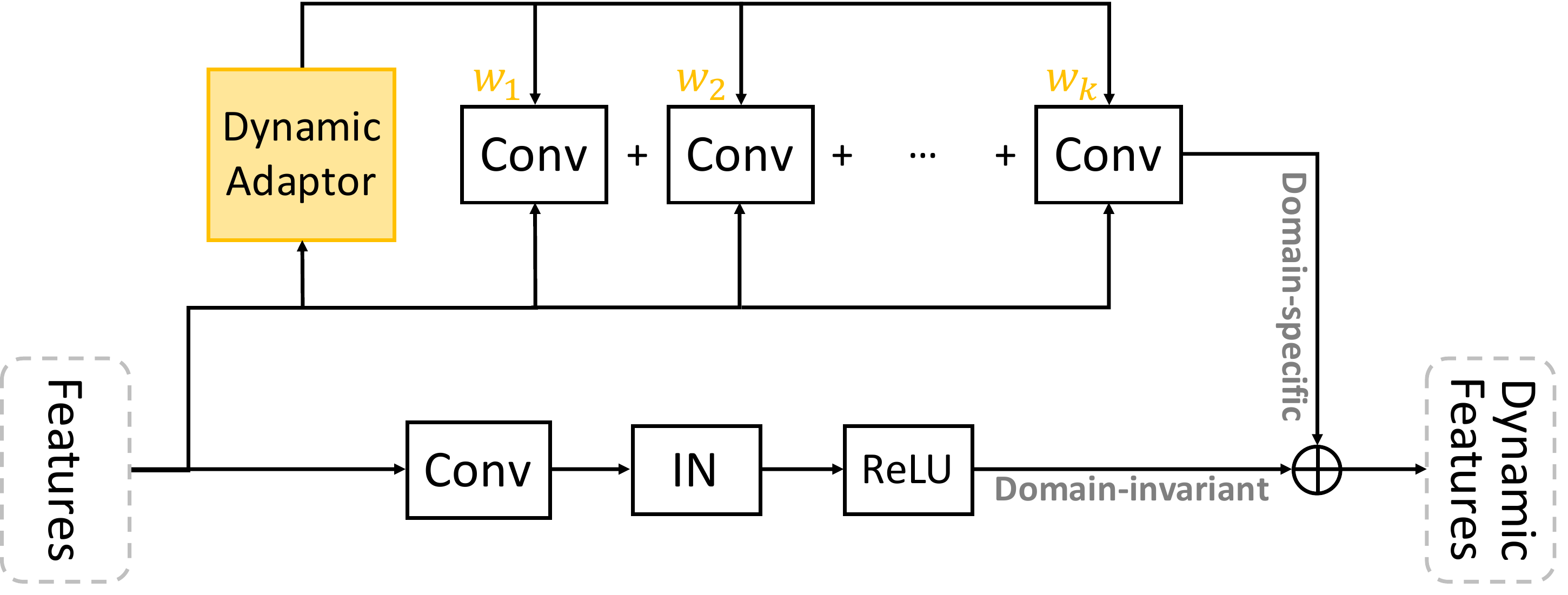}
\end{center}
\vspace{-0.5cm}
\caption{Illustration of dynamic block. }
\label{fig:dyamic_block}
\vspace{-0.3cm}
\end{figure}

Given a feature map $F \in \mathbb{R}^{C \times H \times W}$ of the input image $X$, dynamic block decomposes it to the domain-invariant and domain-specific part at the same time. The whole diagram is shown in Fig. \ref{fig:dyamic_block}. It has two branches. The invariant branch is composed of a common convolution with IN, while the specific one contains multiple convolutions with a dynamic adaptor varying for different samples. 

\textbf{Domain-Invariant Branch.} We combine a common convolution with IN to remove instance-specific characteristics and increase generalization ability of domain-invariant branch. Instead of adopting BN to normalize activations in a mini-batch, IN discards instance-specific information and has demonstrated its capacity in domain generalization \cite{pan2018two, seo2020learning}. Thus, we learn the domain-invariant feature by:
\begin{equation}\label{finv}
F_{inv} = \text{ReLU}(\text{IN}(f^{3\times 3}(F))),
\end{equation}
where \(f^{3\times 3}\) represents a convolution operation with the filter size of $3\times 3$ and ReLU is used as an activation function.

\textbf{Domain-Specific Branch.} Vanilla convolution has a drawback that it has fixed parameters and easily degrades on the unseen domains. To tackle this problem, we introduce a dynamic adaptor to automatically adapt among multiple convolutions. Each sample is regarded as a latent domain and the dynamic adaptor adjusts the network parameters accordingly. 

To be specific, the branch has $K$ convolutions in total and the dynamic adaptor predicts the weights $\mathcal{W} \in \mathbb{R}^{K}$ for each convolution based on the input feature map:
\vspace{-0.1cm}
\begin{equation}\label{adaptor}
\mathcal{W} = d(F),
\vspace{-0.1cm}
\end{equation}
where $d$ is the dynamic adaptor with Pooling-FC-ReLU-FC-Softmax structure. 
The domain-specific feature is the linear combination of K convolutions with dynamic weights:
\vspace{-0.1cm}
\begin{equation}\label{fspec}
F_{spec} = \sum\nolimits_{k=1}^{K} w_k \cdot f^{3\times 3}_{k}(F). 
\vspace{-0.1cm}
\end{equation}

To further increase the diversity of different convolutions, we adopt the Information Maximization (IM) loss $\mathcal{L}_{IM}$ \cite{hu2017learning, liang2020we} to maximize the mutual information between inputs and dynamic weights. It is achieved by entropy minimization $\mathcal{L}_{ent}$ and diversity regularization $\mathcal{L}_{div}$ of the dynamic adaptor predictions: 
\begin{equation}\label{eq:imloss}
	\begin{aligned}
		\mathcal{L}_{ent}(\theta_{\F}, \theta_{\D(X)}) &=    -\mathbb{E}_{\{\mathcal{W}^i\}_{i=1}^{N}} \sum\nolimits_{k=1}^{K} w_k^i \log w_k^i,\\
		\mathcal{L}_{div}(\theta_{\F}, \theta_{\D(X)}) &= \sum\nolimits_{k=1}^{K} \hat{w}_k \log \hat{w}_k \\
			&= D_{KL}(\hat{w}, \frac{1}{K}\mathbf{1}_K) - \log K, \\
		\mathcal{L}_{IM}(\theta_{\F}, \theta_{\D(X)}) &= \mathcal{L}_{ent}(\theta_{\F}, \theta_{\D(X)}) + \mathcal{L}_{div}(\theta_{\F}, \theta_{\D(X)}),
	\end{aligned}
\end{equation}
where $\theta_{\F}$ and $\theta_{\D(X)}$ are the parameters of the feature extractor and dynamic block, $\mathcal{W}^i$ is the K-dimensional weights of the input sample $X_i$ from the dynamic adaptor, $w_k$ is the k-th dimension of $\mathcal{W}$, and $\hat{w} = \mathbb{E}_{\{\mathcal{W}^i\}_{i=1}^{N}} [w^i]$ is the mean weights of $N$ samples. 
As a consequence, the domain-specific branch can dynamically adapt networks in a sample-wise manner and learn more disparate representations via information maximization. 

The final dynamic feature is obtained by simply combing the above two features. 
Our dynamic block ensures domain invariance with IN and adjusts to sample-wise specificity with the dynamic adaptor. Therefore, both domain-invariant and domain-specific features are well represented to facilitate the generalization ability.

\subsection{Meta-Learning Optimization}
\label{ssec:meta}
Single domain generalization is a more realistic but challenging problem for iris PAD. The model needs to generalize to diverse domains with single domain samples. Although data from other domains are not available, we can easily get access to numerous nature images from ImageNet \cite{deng2009imagenet}. Thus, we perturb source images through Fourier transformation to imitate the distribution changes in different domains. 
To empower networks to generalize on the unseen domains, we propose a meta-learning based paradigm. The learning algorithm consists of two phases: meta-train and meta-test. We first conduct normal training based on the single source domain $\mathcal{S}$ in the meta-train phase. Then we learn to generalize to the diverse images from perturbed source domains $\mathcal{S}^+$ in the meta-test phase. The whole process is summarized in Algorithm \ref{alg:meta}. 

\textbf{Meta-Train.} During the meta-train phase, we randomly sample batches on the single source domain $\mathcal{S}$ and conduct classification based on cross-entropy loss: 
\begin{equation}\label{eq:clsloss_train}
	\begin{aligned}
		\mathcal{L}_{Cls(\mathcal{S})}(\theta_\F, \theta_\D, \theta_\C) =& \\
		-\mathbb{E}_{(X^i, Y^i) \sim \mathcal{S}}& \sum\nolimits_{c=1}^{C} Y_c^i \log \C(\D(\F( X^i )))_c,
	\end{aligned}
\end{equation}
where $\theta_\C$ is the parameters of the classifier. The gradient of $\theta_\D$ is calculated with respect to $\mathcal{L}_{Cls}$, and we update the parameters as follows:
\begin{equation}\label{eq:update_train}
	\begin{aligned}
 		{\theta_\D}^\prime = \theta_{\D} - \alpha \nabla_{\theta_\D}\mathcal{L}_{Cls(\mathcal{S})}(\theta_\F, \theta_\D, \theta_\C). 
	\end{aligned}
\end{equation}
 
Besides, the dynamic block $\D$ is optimized using Eq. \ref{eq:imloss}. Note that we only use IM loss in the meta-train phase, therefore, the dynamic adaptor can adaptively combine different convolutions according to the characteristics of each sample in meta-test. 

\textbf{Meta-Test.} In order to imitate the domain shifts from various target domains, we perturb the source images with ImageNet natural images. As the phase component of Fourier spectrum has the semantic-preserving property \cite{oppenheim1981importance, xu2021fourier}, we conduct perturbation through Mixup \cite{zhang2017mixup} in the amplitude spectrum. 
Given a source image $X_s$ and a natural image $X_n$, we first apply Fourier transformation $\mathcal{F}$ with the FFT algorithm \cite{nussbaumer1981fast} and get the corresponding amplitude components $\mathcal{A}(X)$ and phase components $\mathcal{P}(X)$. Then the amplitude spectrum of $\mathcal{A}(X_s)$ is perturbed by $\mathcal{A}(X_n)$ through linear interpolation: 
\begin{equation}\label{eq:perturb}
	\hat{\mathcal{A}}(X_s) = (1-\lambda) \mathcal{A}(X_s) + \lambda \mathcal{A}(X_n), 
\end{equation}
where $\lambda \sim U(0, \eta)$ and $\eta$ controls the strength of the perturbation. The perturbed amplitude spectrum $\hat{\mathcal{A}}(X_s)$ is combined with the original phase spectrum $\mathcal{P}(X_s)$. Finally, we generate the perturbed image $X_{s^+}$ through inverse Fourier transformation $\mathcal{F}^{-1}$.

In the meta-test phase, we first sample batches on the single source domain $\mathcal{S}$ and then perturb them to the extended domains $\mathcal{S}^+$. The meta-test evaluation simulates testing on domains with different distributions, and we encourage our model trained on a single domain to generalize well on the unseen domains. Thus, cross-entropy classification loss is minimized on the perturbed domain $\mathcal{S}^+$:
\begin{equation}\label{eq:clsloss_test}
	\begin{aligned}
		\mathcal{L}_{Cls(\mathcal{S}^+)}(\theta_\F, \theta_\D^\prime, \theta_\C) =& \\
		-\mathbb{E}_{(X^i, Y^i) \sim \mathcal{S}^+}& \sum\nolimits_{c=1}^{C} Y_c^i \log \C(\D(\F( X^i )))_c .
	\end{aligned}
\end{equation}

\textbf{Meta-Optimization.} The meta-train and meta-test are optimized simultaneously. We jointly train the three modules in our network by: 
\vspace{-0.3cm}
\begin{equation}\label{eq:meta}
\footnotesize
	\begin{aligned}
		 \theta_\C \gets \theta_{\C} - \beta  \nabla_{\theta_\C} (& \mathcal{L}_{Cls(\mathcal{S})}(\theta_\F, \theta_\D, \theta_\C)  + \mathcal{L}_{Cls(\mathcal{S}^+)}(\theta_\F, {\theta_\D}^\prime, \theta_\C) ), \\ 
 		\theta_\D \gets  \theta_{\D} - \beta \nabla_{\theta_\D} (& \mathcal{L}_{Cls(\mathcal{S})}(\theta_\F, \theta_\D, \theta_\C) + \mu \mathcal{L}_{IM(\mathcal{S})}(\theta_{\F}, \theta_{\D} )  \\
		  & + \mathcal{L}_{Cls(\mathcal{S}^+)}(\theta_\F, {\theta_\D}^\prime, \theta_\C) ), \\
		  \theta_\F \gets \theta_{\F} - \beta  \nabla_{\theta_\F} (& \mathcal{L}_{Cls(\mathcal{S})}(\theta_\F, \theta_\D, \theta_\C)  + \mu \mathcal{L}_{IM(\mathcal{S})}(\theta_{\F}, \theta_{\D} )  \\
		  & + \mathcal{L}_{Cls(\mathcal{S}^+)}(\theta_\F, {\theta_\D}^\prime, \theta_\C) ) .
	\end{aligned}
\vspace{-0.2cm}
\end{equation}

\begin{algorithm}[t!]
	\caption{Meta Learning of SDDG}
    \label{alg:meta}
	\begin{algorithmic}[1]
        \REQUIRE  \ \\
        \textbf{Input:} A single source domain $\mathcal{S}$ \\
        \textbf{Initialization:} Model parameters $\theta_\F$, $\theta_\D$, $\theta_\C$.  \\
        \textbf{Hyperparameters}: Learning rate $\alpha$ and $\beta$, perturbation strength $\eta$, loss balance $\mu$.
		\WHILE{not done}
            \STATE \textbf{Meta-train:}  Sample batch on the source domain $\mathcal{S}$
            \STATE Evaluate $\mathcal{L}_{Cls(\mathcal{S})}(\theta_\F, \theta_\D, \theta_\C)$ using Eq. \ref{eq:clsloss_train}
            \STATE Compute ${\theta_\D}^\prime$ using Eq. \ref{eq:update_train}
            
            \STATE \textbf{Meta-test:} Sample batch on $\mathcal{S}$ and perturb it to $\mathcal{S}^+$
            \STATE Evaluate $\mathcal{L}_{Cls(\mathcal{S}^+)}(\theta_\F, {\theta_\D}^\prime, \theta_\C)$ using Eq. \ref{eq:clsloss_test}
            
            \STATE \textbf{Meta-optimization:} Update $\theta_\F$, $\theta_\D$, $\theta_\C$ using Eq. \ref{eq:meta}
        \ENDWHILE
	\end{algorithmic}
\end{algorithm}

\begin{table*}[t!]
\vspace{-0.5cm}
 \begin{center}
\begin{threeparttable}
\footnotesize
\caption{Ablation study on different components of SDDG. }
\label{tab:ablation}
  \begin{tabular}{ccc|cc|cc|cc|c}
  \toprule[2pt]
  \multirow{2}{*}{Dynamic Block} & \multirow{2}{*}{$\mathcal{L}_{IM}$} & \multirow{2}{*}{Meta-Learning}  & \multicolumn{2}{c|}{IIITD-WVU} & \multicolumn{2}{c|}{NotreDame} & \multicolumn{2}{c|}{Clarkson}   & \multirow{2}{*}{Average} \\ \cline{4-9}
  & & &  NotreDame     & Clarkson       & IIITD-WVU      & Clarkson      & IIITD-WVU      & NotreDame      &                          \\
  \midrule[1.3pt]
- & - & - & \underline{7.33} & 45.69 & 20.97 & \underline{11.23} & 29.46 & 28.83 & 23.92 \\
\checkmark & - & - & 10.83 & 42.75 & \textbf{15.63} & 14.95 & 26.70 & 17.81 & 21.44 \\
\checkmark & \checkmark & - & 7.94 & 28.89 & \underline{15.84} & 12.22 & 25.95 & 20.39 & 18.54 \\
\checkmark & - & \checkmark & 9.06 & \underline{22.50} & 17.84 & 13.16 & \underline{17.45} & \underline{11.33} & \underline{15.22} \\
\checkmark & \checkmark & \checkmark & \textbf{6.03} & \textbf{19.36} & 16.69 & \textbf{10.20} & \textbf{16.68} & \textbf{8.47} & \textbf{12.90}
\\
  \bottomrule[2pt]
 \end{tabular}
\end{threeparttable}
 \end{center}
\vspace{-0.8cm}
\end{table*}

\begin{table*}[t!]
 \begin{center}
\footnotesize
\caption{Comparison to existing SoTA methods on LivDet-Iris 2017 dataset under cross-dataset settings. $\mathbf{\dagger}$ denotes the reimplemented results.}
\label{tab:sota}
  \begin{tabular}{c|cc|cc|cc|c}
  \toprule[2pt]
  Trained Dataset     & \multicolumn{2}{c|}{IIITD-WVU}    & \multicolumn{2}{c|}{NotreDame}    & \multicolumn{2}{c|}{Clarkson}   & \multirow{2}{*}{Average} \\ \cline{1-7}
  Tested Dataset      & NotreDame       & Clarkson        & IIITD-WVU       & Clarkson        & IIITD-WVU       & NotreDame       &                        \\  \midrule[1.3pt]
  PBS\cite{fang2021iris} & 16.86 & 47.17 & 17.49 & 45.31 & 42.48 & 32.42 & 33.62 \\
 A-PBS\cite{fang2021iris} & 27.61 & 21.99 & \textbf{9.49} & 22.46 & 34.17 & 23.08 & 23.13 \\ 
 D-NetPAD \cite{sharma2020d}{$^\dagger$} & 11.58 & 43.66 & 19.47 & \textbf{8.86} & 27.44 & 17.31 & 21.39 \\ \midrule
 FAM+FMM\cite{li2022few} & \textbf{5.81} & 26.03 & 15.07 & 10.51 & 22.06 & 20.92 & 16.73 \\
 MLDG\cite{li2018learning}{$^\dagger$} & 8.92 & 42.26 & \underline{9.78} & 11.36 & 27.87 & 25.39 & 20.93 \\ \midrule
 MLDG+ & 10.78 & \textbf{17.10} & 18.33 & 18.34 & \underline{19.41} & \underline{10.39} & \underline{15.73} \\
 SDDG & \underline{6.03} & \underline{19.36} & 16.69 & \underline{10.20} & \textbf{16.68} & \textbf{8.47} & \textbf{12.90} \\
  \bottomrule[2pt]
 \end{tabular}
 \end{center}
\vspace{-0.8cm}
\end{table*}

\section{Experiments}
\label{sec:exp}

\subsection{Experimental Setup}
\label{ssec:setup}
\textbf{Datasets.}  We evaluate our method on the LivDet-Iris 2017 dataset \cite{yambay2017livdet}, which consists of 4 different datasets. However, the Warsaw dataset is no longer publicly available, so the experiments are based on the remaining Clarkson, Notre Dame, and IIITD-WVU datasets. Presentation attacks include printed iris images, patterned contact lenses, and printouts of patterned contact lenses. 

\textbf{Implementation Details.} The input image is grayscale and of 200 $\times$ 200 size. Random cropping is performed in the training phase. We adopt ResNet18 as backbone and initial it with ImageNet pre-trained model. The learning rates $\alpha$, $\beta$ are set as 1e-3. Following \cite{xu2021fourier}, the perturbation strength $\eta$ is set to 1. The convolution number $K$ of the dynamic branch and the balance weight $\mu$ of the IM loss are 3 and 1. 

\textbf{Evaluation Metrics.}  For single domain generalization, we use a single dataset for training and conduct evaluations in the rest. Half Total Error Rate (HTER) is employed to measure the performance. 

\subsection{Ablation Study}
\label{ssec:ablation}

\textbf{Effect of different components.} Our method contains three different components: dynamic block, IM loss $\mathcal{L}_{IM}$, and meta-learning. Table \ref{tab:ablation} shows the contribution of each component in the proposed SDDG. We can observe all the components promote the average HTER. Adding dynamic block to the vanilla backbone already boosts the performance both without meta-learning and with it. It verifies that the dynamic representations are effective by leveraging domain-invariant and domain-specific information. Further incorporation of $\mathcal{L}_{IM}$ shows additional improvement with more disparate features. The best and the second best performance of each dataset are almost achieved under meta-learning optimization, which validates the importance of the meta-learning paradigm for Single-DG. The combination of all the components gives best average HTER 12.90\%, outperforming baseline by 11.01\%. 

\textbf{Sensitivity of hyperparameters.}   To validate the significance of the convolution number $K$ in the dynamic branch and balance weight $\mu$ of the IM loss, we conduct sensitivity analysis of the hyperparameters. Fig. \ref{fig:hyperparam} shows the average HTER in single domain generalization by varying $K \in \{2, 3, 4, 5 \}$, $\mu \in \{0, 0.5, 1.0, 1.5, 2.0 \}$. The performance of varying convolution numbers is generally stable varying within 1\%, and best performance is obtained with $K=3$. While introducing $\mathcal{L}_{IM}$ is beneficial for dynamic learning, the moderate $\mu$ provides more promising results. 

\newcommand{\sz}{.5}
\begin{figure}[t]
\vspace{-0.4cm}
\begin{center}
\subfloat[$K$]{\includegraphics[width=\sz\linewidth]{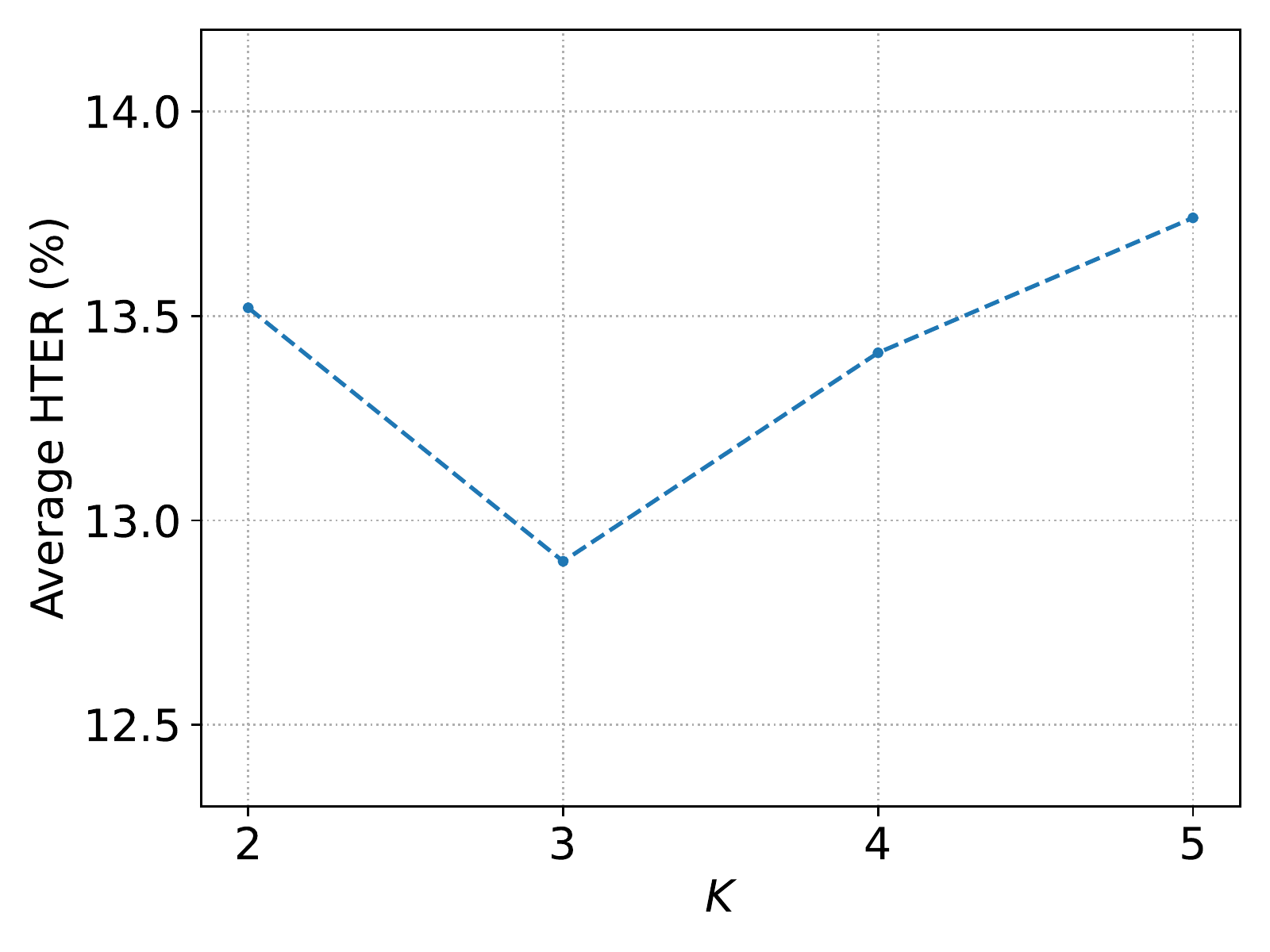} }
\subfloat[$\mu$]{\includegraphics[width=\sz\linewidth]{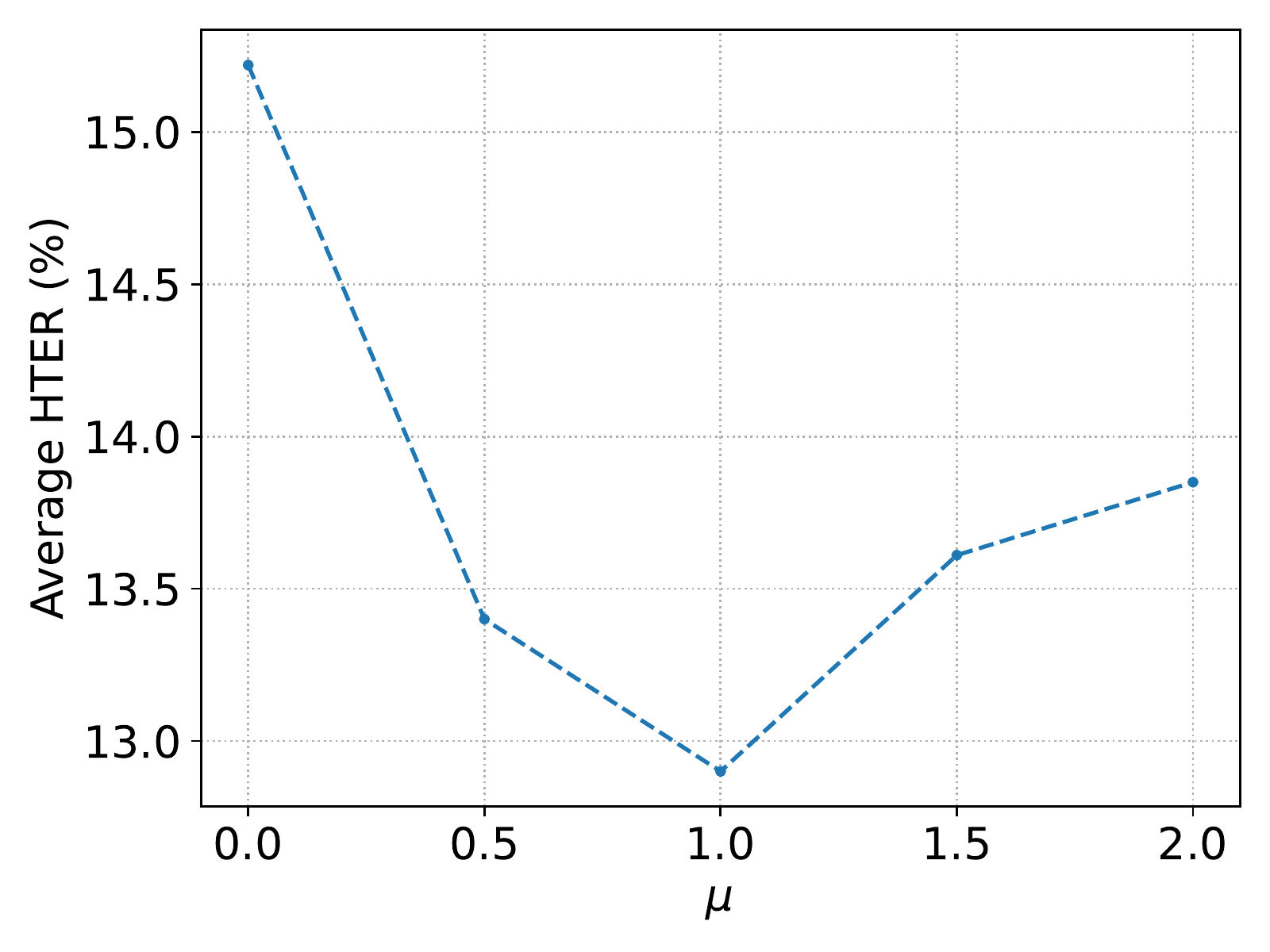}}
\end{center}
\vspace{-0.5cm}
\caption{Evaluation of hyperparameters in Single-DG. }
\label{fig:hyperparam}
\vspace{-0.6cm}
\end{figure}

\subsection{Visualization}
\label{ssec:vis}
In order to investigate how our dynamic block adapts according to samples, we visualize the dynamic weight $\mathcal{W}$ with t-SNE \cite{van2008visualizing}. We can observe that it learns to differentiate attack classes even when the whole network is only trained with binary class labels. It confirms that our dynamic block is capable of adjusting based on the characteristics of each sample even in the challenging Single-DG scenario. 

\begin{figure}[t]
\begin{center}
\includegraphics[width=1.\linewidth]{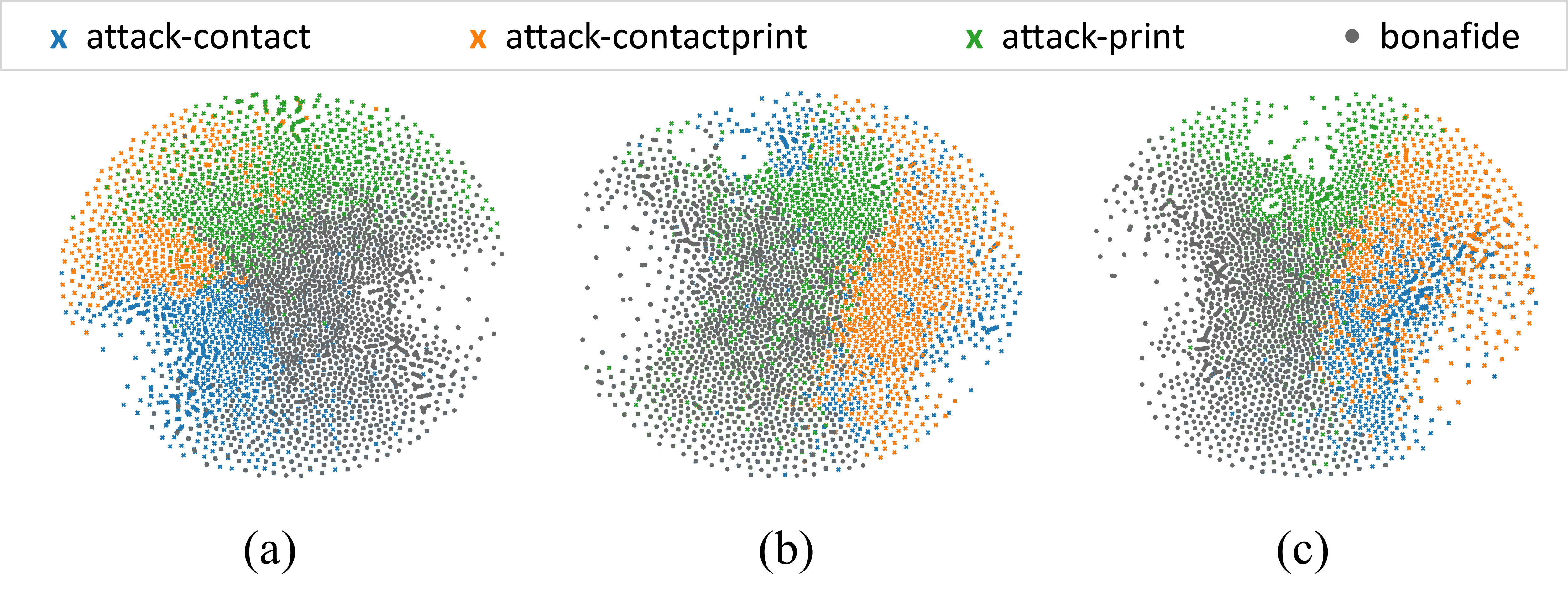}
\end{center}
\vspace{-0.5cm}
\caption{The t-SNE \cite{van2008visualizing} visualization of dynamic weights for all samples from the LivDet-Iris 2017 dataset when SDDG is trained on (a) IIITD-WVU, (b) NotreDame, (c) Clarkson. Note that NotreDame dataset only has contact lens attack, so the corresponding dynamic weights between contact attack and print attack are less discriminative than the others.}
\label{fig:vis}
\vspace{-0.5cm}
\end{figure}

\subsection{Comparison with State-of-the-Art Methods}
\label{ssec:sota}
As shown in Table \ref{tab:sota}, our method outperforms all the state-of-the-art iris PAD methods. PBS \cite{fang2021iris} and A-PBS \cite{fang2021iris} utilize fine-grained pixel-level cues to increase the generalization ability. Although only a single domain is available in the training phase like PBS and A-PBS, SDDG surpasses them by a large margin of 20.72\% and 10.23\%. The purely deep learning-based solution D-NetPAD \cite{sharma2020d} also shows poor performance under cross-dataset settings.
Compared with methods that adopt DA or DG, the proposed method still shows superiority. FAM+FMM \cite{li2022few} method focuses on DA setting and has few target bonafide samples during training. SDDG achieves 3.83\% improvement over FAM+FMM, which demonstrates the effectiveness of our method. Besides, we implement MLDG \cite{li2018learning} to compare the vanilla meta-learning. With a single source domain available, we randomly sample batches on domain $\mathcal{S}$ for both meta-train and meta-test, and meta-optimization is conducted in the classifier. To match the number of parameters increased by the dynamic block, we add an additional convolution to backbone ResNet18 after the feature extractor. The performance is much worse than the proposed SDDG. We further adapt it with perturbed domains $\mathcal{S}^+$ for meta-test, which is referred to as MLDG+. As we can see, MLDG learns to generalize much better with $\mathcal{S}^+$ but is still inferior to the proposed SDDG, which also validates the efficacy of our meta-learning based dynamic framework for Single-DG. 

\section{Conclusion}
\label{sec:conclu}
In this work, we propose a novel single domain dynamic generalization framework for iris presentation detection. Unlike previous methods, SDDG can dynamically adapt based on the characteristics of each sample and exploit domain-invariant and domain-specific features simultaneously. Further incorporation of information maximization loss encourages more disparate representations. Having images perturbed with numerous natural images, the meta-learning paradigm empowers the network to generalize on various unseen domains. Comprehensive experiments validate the effectiveness of the proposed method for single domain generalization.

\vfill\pagebreak

\bibliographystyle{IEEEbib}
\bibliography{refs}

\end{document}